\documentclass[10pt,twocolumn,letterpaper]{article}

\usepackage{cvpr}              

\usepackage{graphicx}
\usepackage{amsmath}
\usepackage{amssymb}
\usepackage{booktabs}
\usepackage{xcolor}
\usepackage{wrapfig}

\usepackage[pagebackref,breaklinks,colorlinks]{hyperref}

\usepackage[capitalize]{cleveref}
\crefname{section}{Sec.}{Secs.}
\Crefname{section}{Section}{Sections}
\Crefname{table}{Table}{Tables}
\crefname{table}{Tab.}{Tabs.}

\begin{document}

\title{Localized Latent Updates for Fine-Tuning Vision-Language Models}

\author{%
{\parbox{\textwidth}{\centering \hspace{\stretch{2}} Moritz Ibing\hspace{\stretch{1.5}}Isaak Lim\hspace{\stretch{1.5}}Leif Kobbelt\hspace{\stretch{2}} }} \\
  Visual Computing Institute, RWTH Aachen University
}
\maketitle

\begin{abstract}
Although massive pre-trained vision-language models like CLIP show impressive generalization capabilities for many tasks,
still it often remains necessary to fine-tune them for improved performance on specific datasets.
When doing so, it is desirable that updating the model is fast and that the model does not lose its capabilities on data outside of the dataset,
as is often the case with classical fine-tuning approaches.
In this work we suggest a lightweight adapter, that only updates the models predictions close to seen datapoints.
We demonstrate the effectiveness and speed of this relatively simple approach in the context of few-shot learning,
where our results both on classes seen and unseen during training are comparable with or improve on the state of the art.
\end{abstract}

\section{Introduction}
Much of the success of deep learning when it comes to vision tasks, such as classification, object detection or segmentation, is due to ever bigger models trained on increasingly large quantity of data. 

A popular approach to make use of immense sources of uncurated data in the form of images with textual descriptions are vision-language models \cite{radford2021learning, jia2021scaling}. Here both image and description are individually mapped into a joint embedding space. This embedding is optimized, so that matching pairs are close, and the distance between all other pairs large. A model trained in this fashion can be used for zero-shot classification, as the language model can deal with every conceivable class by embedding a textual description (\eg "a picture of [CLASS]"). 

Even though these models can be applied to all kinds of classification tasks, their performance sometimes is sub-optimal. This might be the case if used on a dataset with specific characteristics, that differ from the original training set, \eg if the task is to recognize the action performed in an image, even though during training the model only saw generic objects. In these cases a common technique is to fine-tune the pre-trained model for the task at hand. However, updating the complete model is quite expensive (as the employed models are large). There are two solutions proposed in the literature to tackle this problem. 
One is prompt-learning \cite{zhou2022learning, zhou2022conditional, zhu2022prompt} where the context around the class ("a picture of" in the last example) is optimized for a specific dataset instead of hand crafted. 
The other option is to use adapters\cite{gao2021clip, zhang2021tip}, light-weight models (usually small MLPs) that modify the embedding produced by either the visual or language model (or both), thus updating the predictions without the need to update the original networks parameters.

Both these approaches however still have the problem, that even though the performance is improved for the specific domain and task the fine-tuning was done for, this comes at the cost of a decrease in performance on other tasks/domains compared to the original model \cite{goodfellow2013empirical}.

The goal of this work is to reap the benefits of fine-tuning on a specific task, without losing the generalization ability of the original model. Work in this direction has already been done in the form of CoCoOp \cite{zhou2022conditional}, where prompt-learning is employed, but the context is not fine-tuned on a specific dataset, but instead a suitable context is predicted from the image to be classified. Another approach using prompt-learning is ProGrad \cite{zhu2022prompt} where the context update is restricted in order not to lose information from the pre-training stage. Although both methods decrease the performance loss in the zero-shot setting, they still do not reach the abilities of the original model. 

In contrast we choose a simple method based on adapters. The idea is to only update the embedding where we actually have training data and leave it unchanged everywhere else, thus retaining the original predictions of the model, where we cannot improve on them. Furthermore, even where we have data we want to change the embedding as little as possible, to allow sensible interpolation between fine-tuned and original embedding. This approach is extremely lightweight, as we only need to tune a small amount of parameters, and back-propagating through the original model is not necessary. Still we show an improvement in performance compared to the previous state of the art.

\section{Related Work}
\paragraph{Zero-Shot Learning}
Zero-shot learning describes a setting, where the set of classes at training and during testing are disjoint or at least not identical \cite{chao2016empirical}, thus the relation between classes and images belonging to that class needs to be learned indirectly. There are numerous works of research in this area, so we instead refer to \cite{xian2017zero, wang2019survey} for an overview. A common approach to tackle this task is to relate pre-trained image and class embeddings \cite{akata2015label, akata2015evaluation, frome2013devise}. This is conceptually very close to vision-language models, another popular framework that can be applied for zero-shot learning.

\paragraph{Vision-Language Models}
The term vision-language model describes networks that learn an alignment between images and text in a joint embedding space. Today these models are usually based on contrastive learning, which has been popularized for pre-training of image models \cite{chen2020simple, he2020momentum, henaff2020data} and aims to maximize the distance between similar and dissimilar instances in the embedding space. Currently one of the most popular vision-language models is CLIP \cite{radford2021learning}, which we use in all our experiments. It uses a Transformer \cite{vaswani2017attention} as text encoder and a ResNet \cite{he2016deep} or ViT \cite{dosovitskiy2020vit} as image encoder. ALIGN \cite{jia2021scaling} is a similar approach, whereas DeCLIP \cite{li2022supervision} tries to improve the training procedure in order reach the same performance with less data.

\paragraph{Fine-Tuning}
When it comes to fine-tuning a pre-trained vision-language model, there are broadly three types of approaches in the literature. It is possible to fine-tune the entire model, but afterwards interpolate between the original and updated weights, to counteract overfitting (WiSE-FT \cite{wortsman2022robust}).
Alternatively, not the model itself is trained, but only an adapter, that is applied onto the embedding space (CLIP-Adapter \cite{gao2021clip}). This is the approach we choose as well. Instead of learning this adapter, it can be extracted from the fine-tuning dataset (TIP-Adapter \cite{zhang2021tip}). As this requires data for every class it is evaluated on, it is however not suitable for zero-shot learning. 
The last approach is called prompt engineering. Here the context of the text embedding is optimized for performance on the training set (CoOp \cite{zhou2022learning}). This learning can be restricted in order not to decrease the loss of information from the pre-training stage (ProGrad \cite{zhu2022prompt}). Another option is to predict the (textual) context for each image (CoCoOp \cite{zhou2022conditional}). This mitigates overfitting on the train set and thus is better at retaining zero shot ability on unseen classes, but comes at the cost of an increased training time.

\newpage

\section{Method}

Before introducing our approach in more detail, we will give a short overview on how vision-language models work in general on the example of CLIP, which is used in all our experiments.

\subsection{Vision-Language Models}
Vision-Language models consist of two networks: an image encoder $f_I$ and a text encoder $f_T$. Their exact implementation is of no interest to us in this context. All we need to know, is that these models embed an image or a text respectively to a (normalized) feature vector of the same dimension. The cosine distance between an embedded image and text should then correspond to their similarity \ie how well the text describes the image.

During training, we are given a batch of $n$ images $X$ and their textual descriptions $Y$. We make the simplifying assumption that each text is a perfect description of the corresponding image and all other texts are completely unrelated. Thus, we want to minimize the cosine distance between embeddings of matching image/text pairs $f_I(x_i)$,$f_T(y_i)$ and maximize the distance between all other pairs within the batch $f_I(x_i)$,$f_T(y_j)$ with $i \neq j$. 

Another view would be to regard the cosine distance as the likelihood, that a given text $y$ describes the corresponding image $x$, or vice versa. We can compute the normalized probabilities as:
\begin{equation}
    p(y|x) = \frac{\text{exp}(f_I(x)^Tf_T(y)/\tau)}{\sum_{i=1}^{n}\text{exp}(f_I(x)^Tf_T(y_i)/\tau)}
\end{equation}
where $\tau$ is a learned temperature parameter. As we assume the embeddings to be normalized, the dot product is equivalent to the cosine similarity. The probability $p(x|y)$ only differs in the normalization.

In this view it makes now sense to maximize the probability for the correct pairs, for which we can use the Cross Entropy loss. As we want to maximize the probabilities in both directions the loss is given as:
\begin{equation}
    L = -\frac{1}{N}\sum_{i=1}^N(\text{log}(p(x_i|y_i)) + \text{log}(p(y_i|x_i)))/2
\end{equation}

\paragraph{Zero-Shot Classification}
This approach leads to a semantically meaningful embedding of both images and text, that can be used for downstream tasks. Alternatively, we can use it directly for zero-shot classification.
For this we embed a text for each class (\eg a picture of [CLASS]) to a feature vector $f_T(y_k)$. Then, to classify a given image, we embed it and compute the distance between its feature vector $f_I(x)$ and all class embeddings. The class probability for a class $k$ is then given similarly as before as $p(y_k|x)$.

\subsection{Local Linear Updates}
\begin{figure}[t]
    \centering
    \includegraphics[width=1.0\linewidth]{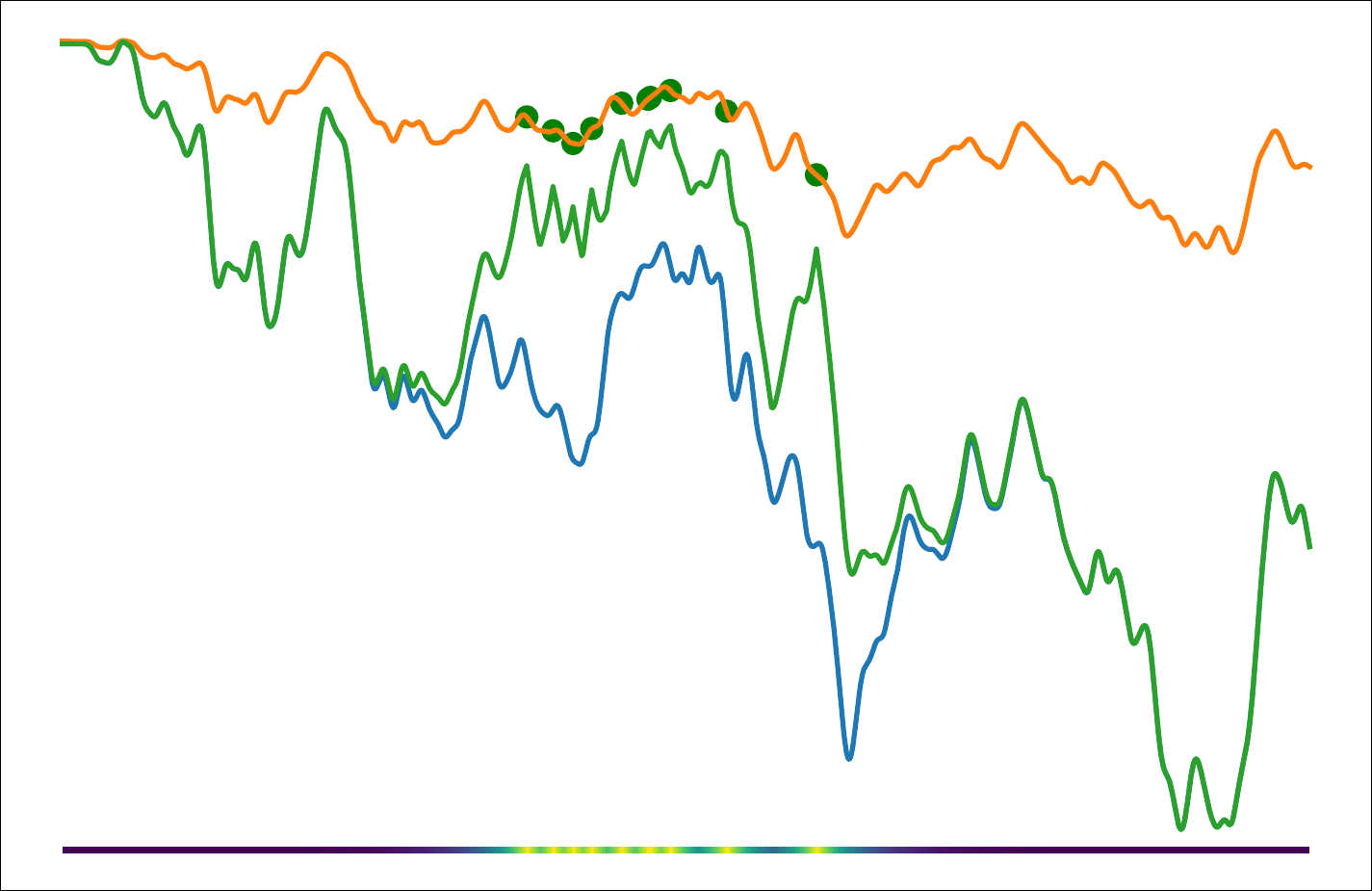}
    \hfill
    \caption{The basic idea of our approach is to interpolate between the original networks output (blue) and the newly trained adapter (orange) to obtain the final function (green). The interpolation weight (bottom line) is based on the distance to the samples we fine-tuned on (green points). }
    \label{fig:overview}
\end{figure}

A big benefit of vision-language models is their generalization capability. As they are usually trained on immense datasets, they tend to show great zero-shot capabilities even on unseen domains. However, they are not necessarily well tuned for small specific datasets, that were not well represented in the original training data.

Although we could fine-tune the networks on this specific set, this would come at the cost of the models ability to generalize and be quite expensive. Instead we add additional functions on the output of the model $g_I \circ f_I$ and $g_T \circ f_T$ respectively and optimize only those. This is much cheaper, as $g$ has much less parameters (we will omit the subscript for both $f$ and $g$ if both the image and textual models are meant). Furthermore, as $g$ is applied onto the output of $f$ we do not even have to back-propagate through the big models. We could even precompute $f$ on the given dataset to save further computation cost. In our case we use distinct networks for $g_I$ and $g_T$ but is possible to use the same function as well ($g_I = g_T$) as they are applied on the same domain. 

The training works slightly differently than before. As we now do not have unique image/text pairs, but instead images with their classes, we leave out one half of the loss, leading to the classical Cross Entropy loss:
\begin{equation}
    L = -\frac{1}{N}\sum_{i=1}^N\text{log}(p(y_i|x_i))
\end{equation}

A similar idea was presented in CLIP Adapter\cite{gao2021clip} (Here only $g_I$ is used). 
This approach significantly reduces the cost of fine-tuning, but does not solve the problem of loosing capabilities on the previous training dataset while overfitting on the new one. 

\paragraph{Local Interpolation}
To lessen the overfitting WISE-FT\cite{wortsman2022robust} interpolates the weights between the original model and a fine-tuned version.
With a similar aim in CLIP Adapter \cite{gao2021clip} the results of $f$ and $g$ are interpolated:
\begin{equation}
    \alpha (g \circ f) + (1-\alpha)f
\end{equation}
Here $\alpha$ is a global parameter. It would be more sensible to localize this interpolation to the area of the feature space, where we obtained new data. Only there do we actually have information onto how to sensibly update the embeddings. 
As $g$ is a global function but we only supervise it at our training samples, it is unlikely that it represents a sensible modification away from these samples.
Thus, in our case $\alpha$ is not a global parameter, but a function,
\begin{equation}
\alpha(x,D) = \beta \cdot \text{max}_{d \in D}(\text{exp}(-\gamma(1 -x^Td)))
\end{equation}
where $D$ is the set of datapoints we fine-tuned on and $\beta$ is a global parameter similar to how $\alpha$ was defined previously. $\gamma$ concentrates the focus of the interpolation mask and thus influences in what range our updated embedding should be applied. If we would let it go towards zero, we would have a global $\alpha$ parameter, similarly to CLIP Adapter. On the other hand, if we let it go towards infinity, we would only update exactly the images and classes on which we fine-tuned. 

Whenever an image is close to one already seen during training, we thus use our updated features, otherwise we utilize the general knowledge of the pre-trained model. Note that we do this separately for the text and image encoder, so there are separate sets $D_{text}$ and $D_{image}$. 

\paragraph{Clustering}
This approach requires us to save the feature vectors of all datapoints seen during fine-tuning. This is not a problem, as long as the dataset used is indeed very small. If this however is not the case, we cluster the feature vectors to find sensible representatives. For this we use agglomerative clustering, where we start by regarding each datapoint as an individual cluster and then iteratively merge pairs based on the maximum distance between their members until we have reached the desired number of clusters. Each cluster needs a position, which is computed as the (normalized) mean of all its members. 

We chose this approach, as it does not make any assumptions about cluster shapes nor does it require a sensible initialization. Furthermore, we expect the number of clusters to be in a similar order of magnitude as the number of datapoints, so we do not need many merge operations. However, as we show in the ablation (Sec.\ref{pg:ablation}), clustering only has a small effect on the performance and is mainly used to bound the memory consumption. Thus, the exact clustering algorithm is not likely to make much of a difference either.

\paragraph{Identity Regularization}
So far, we have restricted the region, where we change the feature space, but not the magnitude of the update, which can become arbitrarily large. It is however desirable, that the update is as small as possible, while minimizing the training loss. As we assume the original pre-trained features to already be useful, we want to stay as close to them as possible in order to retain generality. 
Furthermore the interpolation between $f$ and $g \circ f$ should result in sensible embeddings, which is more likely the case, if they are close to each other. In other words, $g$ should stay as close to the identity as possible.

This is easy to enforce, if we simply choose $g$ as an affine function $g = Wx + b$. In this case our regularization takes the form:
\begin{equation}
    \lambda(||W - I||_2 + ||b||_2)
\end{equation}
where $I$ is the identity matrix and $\lambda$ a weighting parameter. 

Of course we could choose a more complex function and regularize $g$ to stay close to the identity at a set of sample points. However, a dense sampling of the embedding space is infeasible, and we are interested in retaining this property wherever the interpolation weight $\alpha$ is non-zero.

Furthermore, we initialize $g$ as the identitiy function, which is trivial for affine functions, but not for non-linear MLPs.

\begin{figure*}[t]
    \includegraphics[width=1.0\linewidth]{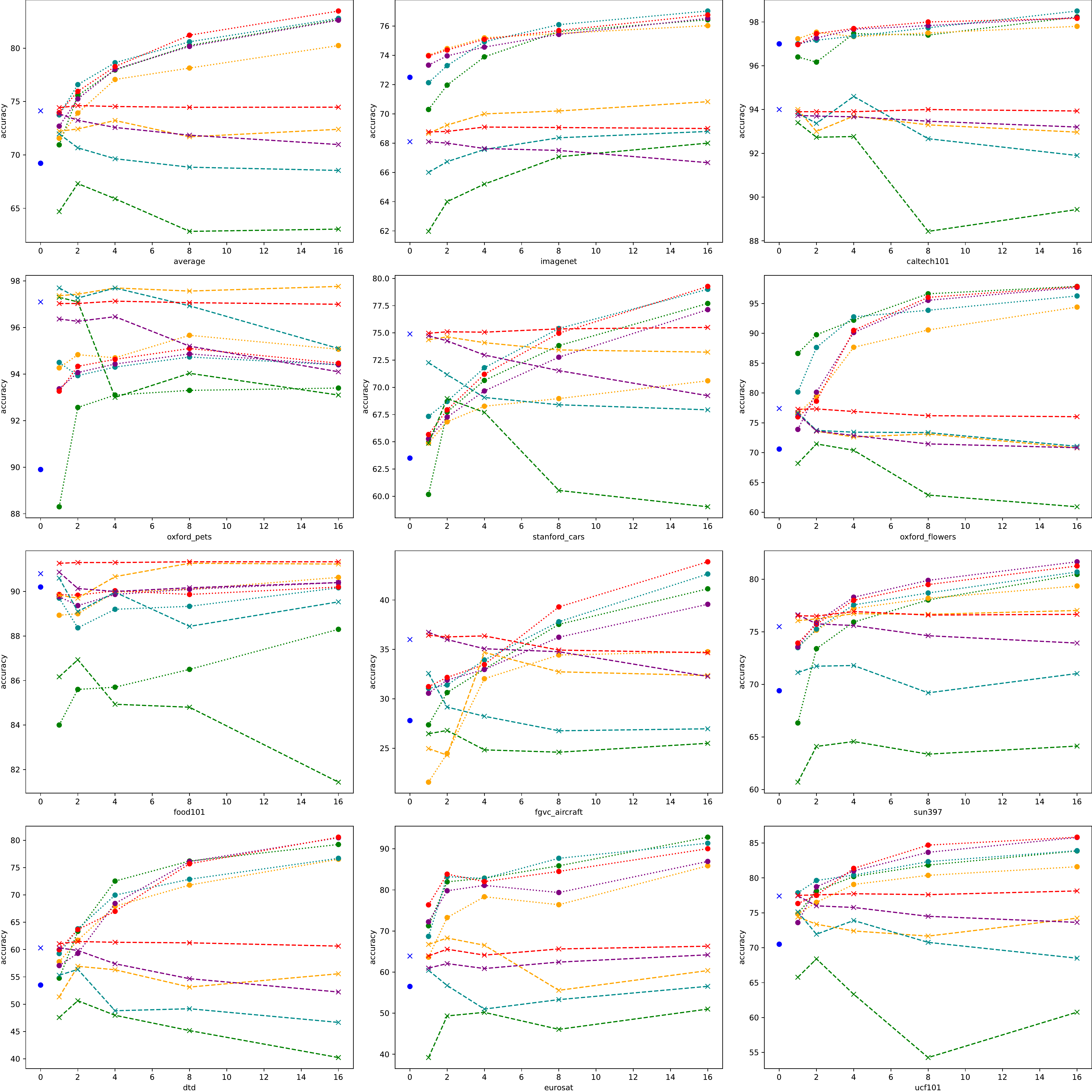}
    \caption{Comparison in the intra-class generalization setting. We compare our approach (red) vs. CoOp (green), CoCoOp (yellow), ProGrad (cyan) and CLIP Adapter (purple). Zero-shot CLIP is shown as a baseline in blue. circles mark the base classes and x the unseen new classes.} 
    \label{fig:base2new}
\end{figure*}

\section{Evaluation}

\begin{table*}
    \begin{subtable}[h]{0.33\textwidth}
        \begin{tabular}{ccc|c}
\toprule
        & Base  & New   & H     \\
\midrule
CLIP            & 69.34 & 74.22 & 71.70 \\
CoOp            & 82.69 & 63.22 & 71.66 \\
CoCoOp          & 80.47 & 71.69 & 75.83 \\
ProGrad         & 82.79 & 68.55 & 74.46 \\
CLIP Ada.       & 82.62 & 70.97 & 76.02 \\
LLU             & \textbf{83.48} & \textbf{74.47} & \textbf{78.46} \\
\bottomrule                          
\end{tabular}
        \caption{Average over 11 datasets}
    \end{subtable}
    \hfill
    \begin{subtable}[h]{0.33\textwidth}
        \begin{tabular}{ccc|c}
\toprule
        & Base  & New   & H     \\
\midrule
CLIP        & 72.43 & 68.14 & 70.22 \\
CoOp        & 76.47 & 67.88 & 71.92 \\
CoCoOp      & 75.98 & \textbf{70.43} & \textbf{73.10} \\
ProGrad     & \textbf{77.03} & 68.8 & 72.68 \\
CLIP Ada.   & 76.53 & 66.67 & 71.26 \\
LLU         & 76.77 & 69.00 & 72.68 \\
\bottomrule                          
\end{tabular}
        \caption{ImageNet}
    \end{subtable}
    \hfill
    \begin{subtable}[h]{0.33\textwidth}
        \begin{tabular}{ccc|c}
\toprule
        & Base  & New   & H     \\
\midrule
CLIP        & 96.84 & \textbf{94.00} & 95.40 \\
CoOp        & 98.00 & 89.81 & 93.73 \\
CoCoOp      & 97.96 & 93.81 & 95.84 \\
ProGrad     & \textbf{98.50} & 91.90 & 95.09 \\
CLIP Ada.   & 98.20 & 93.20 & 95.63 \\
LLU         & 98.17 & 93.93 & \textbf{96.00} \\
\bottomrule                          
\end{tabular}
        \caption{Caltech101}
    \end{subtable}

    \begin{subtable}[h]{0.33\textwidth}
        \begin{tabular}{ccc|c}
\toprule
        & Base  & New   & H     \\
\midrule
CLIP        & 91.17 & 97.26 & 94.12 \\
CoOp        & 93.67 & 95.29 & 94.47 \\
CoCoOp      & \textbf{95.20} & \textbf{97.69} & \textbf{96.43} \\
ProGrad     & 94.40 & 95.10 & 94.75 \\
CLIP Ada.   & 94.40 & 94.10 & 94.25 \\
LLU         & 94.47 & 97.00 & 95.72 \\
\bottomrule                          
\end{tabular}
        \caption{OxfordPets}
    \end{subtable}
    \hfill
    \begin{subtable}[h]{0.33\textwidth}
        \begin{tabular}{ccc|c}
\toprule
        & Base  & New   & H     \\
\midrule
CLIP        & 63.37 & 74.89 & 68.65 \\
CoOp        & 78.12 & 60.40 & 68.13 \\
CoCoOp      & 70.49 & 73.59 & 72.01 \\
ProGrad     & 79.00 & 67.93 & 73.05 \\
CLIP Ada.   & 77.13 & 69.23 & 72.97 \\
LLU         & \textbf{79.27} & \textbf{75.50} & \textbf{77.34} \\
\bottomrule                          
\end{tabular}
        \caption{StanfordCars}
    \end{subtable}
    \hfill
    \begin{subtable}[h]{0.33\textwidth}
        \begin{tabular}{ccc|c}
\toprule
        & Base  & New   & H     \\
\midrule
CLIP        & 72.08 & \textbf{77.80} & 74.83 \\
CoOp        & 97.60 & 59.67 & 74.06 \\
CoCoOp      & 94.87 & 71.75 & 81.71 \\
ProGrad     & 96.27 & 71.07 & 81.77 \\
CLIP Ada.   & 97.70 & 70.83 & 82.13 \\
LLU         & \textbf{97.83} & 76.03 & \textbf{85.57} \\
\bottomrule                          
\end{tabular}
        \caption{Flowers102}
    \end{subtable}

    \begin{subtable}[h]{0.33\textwidth}
        \begin{tabular}{ccc|c}
\toprule
        & Base  & New   & H     \\
\midrule
CLIP        & 90.10 & 91.22 & 90.66 \\
CoOp        & 88.33 & 82.26 & 85.19 \\
CoCoOp      & \textbf{90.70} & 91.29 & \textbf{90.99} \\
ProGrad     & 90.17 & 89.53 & 89.85 \\
CLIP Ada.   & 90.40 & 90.40 & 90.40 \\
LLU         & 90.20 & \textbf{91.33} & 90.76 \\
\bottomrule                          
\end{tabular}
        \caption{Food101}
    \end{subtable}
    \hfill
    \begin{subtable}[h]{0.33\textwidth}
        \begin{tabular}{ccc|c}
\toprule
        & Base  & New   & H     \\
\midrule
CLIP        & 27.19 & \textbf{36.29} & 31.09 \\
CoOp        & 40.44 & 22.30 & 28.75 \\
CoCoOp      & 33.41 & 23.71 & 27.74 \\
ProGrad     & 42.63 & 26.97 & 33.04 \\
CLIP Ada.   & 39.57 & 32.27 & 35.55 \\
LLU         & \textbf{43.87} & 34.67 & \textbf{38.72} \\
\bottomrule                          
\end{tabular}
        \caption{FGVCAircraft}
    \end{subtable}
    \hfill
    \begin{subtable}[h]{0.33\textwidth}
        \begin{tabular}{ccc|c}
\toprule
        & Base  & New   & H     \\
\midrule
CLIP        & 69.36 & 75.35 & 72.23 \\
CoOp        & 80.60 & 65.89 & 72.51 \\
CoCoOp      & 79.74 & \textbf{76.86} & 78.27 \\
ProGrad     & 80.70 & 71.03 & 75.56 \\
CLIP Ada.   & \textbf{81.67} & 73.93 & 77.61 \\
LLU         & 81.27 & 76.67 & \textbf{78.90} \\
\bottomrule                          
\end{tabular}
        \caption{SUN397}
    \end{subtable}

    \begin{subtable}[h]{0.33\textwidth}
        \begin{tabular}{ccc|c}
\toprule
        & Base  & New   & H     \\
\midrule
CLIP        & 53.24 & 59.90 & 56.37 \\
CoOp        & 79.44 & 41.18 & 54.24 \\
CoCoOp      & 77.01 & 56.00 & 64.85 \\
ProGrad     & 76.70 & 46.67 & 58.03 \\
CLIP Ada.   & 80.47 & 52.23 & 63.35 \\
LLU         & \textbf{80.56} & \textbf{60.63} & \textbf{69.19} \\
\bottomrule                          
\end{tabular}
        \caption{DTD}
    \end{subtable}
    \hfill
    \begin{subtable}[h]{0.33\textwidth}
        \begin{tabular}{ccc|c}
\toprule
        & Base  & New   & H     \\
\midrule
CLIP        & 56.48 & 64.05 & 60.03 \\
CoOp        & \textbf{92.19} & 54.74 & 68.69 \\
CoCoOp      & 87.49 & 60.04 & 71.21 \\
ProGrad     & 91.37 & 56.53 & 69.85 \\
CLIP Ada.   & 86.93 & 64.20 & 73.86 \\
LLU         & 90.33 & \textbf{66.30} & \textbf{76.37} \\
\bottomrule                          
\end{tabular}
        \caption{EuroSAT}
    \end{subtable}
    \hfill
    \begin{subtable}[h]{0.33\textwidth}
        \begin{tabular}{ccc|c}
\toprule
        & Base  & New   & H     \\
\midrule
CLIP        & 70.53 & 77.50 & 73.85 \\
CoOp        & 84.69 & 56.05 & 67.46 \\
CoCoOp      & 82.33 & 73.45 & 77.64 \\
ProGrad     & 83.90 & 68.50 & 75.42 \\
CLIP Ada.   & 85.80 & 73.63 & 79.25 \\
LLU         & \textbf{85.83} & \textbf{78.13} & \textbf{81.80} \\
\bottomrule                          
\end{tabular}
        \caption{UCF101}
    \end{subtable}
    \caption{Comparison in the intra-class generalization setting. All methods except for CLIP (CoOp, CoCoOp, ProGrad, CLIP Adapter and our method LLU (Localized Linear Updates)) are trained on the base classes with 16 shots. H denotes the harmonic mean.}
    \label{tab:inner_dataset}
\end{table*}

\begin{table*}
    \centering
    \begin{tabular}{c|c|cccccccccc|c}
\toprule
        & Source    & \multicolumn{10}{c}{Target} \\
\midrule
        & \rotatebox[origin=c]{90}{ImageNet} & \rotatebox[origin=c]{90}{Caltech101} & \rotatebox[origin=c]{90}{OxfordPets} & \rotatebox[origin=c]{90}{StanfordCars} & \rotatebox[origin=c]{90}{Flowers102} & \rotatebox[origin=c]{90}{Food101} & \rotatebox[origin=c]{90}{FGCVAircraft} & \rotatebox[origin=c]{90}{SUN397} & \rotatebox[origin=c]{90}{DTD} & \rotatebox[origin=c]{90}{EuroSAT} & \rotatebox[origin=c]{90}{UCF101} & \rotatebox[origin=c]{90}{Average} \\
\midrule
CLIP            & 62.57 & 90.17 & 89.10 & 65.30 & 71.40 & \textbf{86.10} & 24.80 & 57.53 & 44.40 & \textbf{47.80} & 66.70 & 64.33 \\
CoOp            & 71.51 & 93.70 & 89.14 & 64.51 & 68.71 & 85.30 & 18.47 & 64.15 & 41.92 & 46.39 & 66.55 & 63.88 \\
CoCoOp          & 71.02 & \textbf{94.43} & \textbf{90.14} & 65.32 & \textbf{71.88} & 86.06 & 22.94 & \textbf{67.36} & \textbf{45.73} & 45.37 & \textbf{68.21} & \textbf{65.74} \\
ProGrad         & 72.00 & 92.67 & 89.73 & 64.00 & 68.37 & 85.27 & 20.30 & 64.60 & 43.07 & 44.53 & 65.20 & 63.78 \\
CLIP Adapter    & 71.77 & 92.17 & 86.47 & 60.50 & 67.63 & 82.53 & 22.90 & 62.77 & 42.23 & 47.67 & 63.37 & 62.82 \\
LLU             & \textbf{72.13} & 92.00 & 89.10 & \textbf{65.37} & 71.23 & \textbf{86.10} & \textbf{24.87} & 64.93 & 44.63 & 47.77 & 67.10 & 65.31 \\
\bottomrule                          
\end{tabular}

    \caption{Comparison for cross dataset generalization capability. All approaches are trained on ImageNet (16 shots) and then evaluated on all 11 datasets.}
    \label{tab:cross_dataset}
\end{table*}

\begin{table*}
    \centering
    \begin{tabular}{c|c|cccc}
\toprule
        & Source    & \multicolumn{4}{c}{Target} \\
\midrule
        & ImageNet & ImageNetV2 & ImageNet-Sketch & ImageNet-A & ImageNet-R \\
\midrule
CLIP            & 66.73 & 60.83 & 46.15 & 47.77 & 73.96  \\
CoOp            & 71.51 & 64.20 & 47.99 & 49.71 & 75.21  \\
CoCoOp          & 71.02 & 64.07 & \textbf{48.75} & \textbf{50.63} & \textbf{76.18}  \\
ProGrad         & 72.00 & \textbf{64.70} & 48.37 & 49.73 & 75.57  \\
CLIP Adapter    & 71.77 & 63.97 & 46.27 & 47.80 & 72.10  \\
LLU             & \textbf{72.13} & 64.53 & 47.17 & 48.87 & 74.30  \\
\bottomrule                          
\end{tabular}
    \caption{Comparison for domain generalization capability. All approaches are trained on the standard version of ImageNet (16 shots) and then evaluated on 4 different types of domain shift.}
    \label{tab:domain}
\end{table*}

For our evaluation we follow CoCoOp\cite{zhou2022conditional}, where three problem settings are investigated:
\begin{enumerate}
    \item Generalization to new classes \emph{within} a given dataset. 
    \item Generalization to new datasets after fine-tuning. 
    \item Generalization to domain-shift.
\end{enumerate}
Before presenting the conclusions, we will introduce the used datasets, and explain the training procedure. 

\paragraph{Datasets}
Similar to CoCoOp, we follow CoOp \cite{zhou2022learning} in the choice of datasets used in evaluation. To be precise we use 11 datasets that cover a wide range of tasks: ImageNet \cite{deng2009imagenet} and Caltech101 \cite{fei2004learning} for generic object classification, OxfordPets \cite{parkhi2012cats}, StanfordCars \cite{krause20133d}, Flowers102 \cite{nilsback2008automated}, Food101 \cite{bossard2014food} and FGVCAircraft for more specific object classification, \cite{maji2013fine}, SUN397 \cite{xiao2010sun}, DTD \cite{cimpoi2014describing}, EuroSAT \cite{helber2019eurosat} and UCF101 \cite{soomro2012ucf101} for a diverse set of tasks. Furthermore, to evaluate domain generalization we regard ImageNet as source and four different versions under different types of domain shift as target. The four datasets are: ImageNetV2 \cite{recht2019imagenet}, ImageNet-Sketch \cite{wang2019learning}, ImageNet-A \cite{hendrycks2021natural} and ImageNet-R \cite{hendrycks2021many}. 

The set of images for few-shot training are randomly sampled for each dataset, while using the original test set for testing. For approaches that need training we average the results over three runs.

\paragraph{Training}
Our implementation is based on the published code of CoOp. We use the same learning rate and number of epochs as they do. Following CoCoOp we use ViT-B/16 as the vision backbone of CLIP. As ProGrad has been evaluated on a different backbone, we retrained it for a fair comparison.
Note that both CoOp and CoCoOp have a context length of 4 initialized as the prompt: "a photo of a", whereas for CLIP Adapter and us the context is class dependent and ProGrad has a context length of 16 with a class-dependent initialization. If not stated otherwise we choose the parameters of our approach as $\beta = 0.5$, $\gamma = 20$, $\lambda = 1e3$ and the number of clusters as 512.

\subsection{Base to New Generalization}
On each dataset, the classes are split equally into a set of "base" classes on which the adapter is trained and unseen "new" classes,
where we only evaluate. Thus, no matter how many shots are given for the training, on the new classes we will always do zero shot inference.
We show results for different numbers of shots in Figure \ref{fig:base2new} and report exact values in Table \ref{tab:inner_dataset}.
Here we also give the harmonic mean between the evaluation on base and new classes for an easier comparison of the approaches regarding their respective trade-offs.

As can be seen, on the 16 shot evaluation our approach outperforms all other methods on 8 out of 11 datasets, when regarding the harmonic mean. Here we have on average an improvement of almost 3 percentage points to CoCoOp (the next best method). Furthermore, (on average) our localized adapter beats all other methods regarding new classes independent of the number of shots and is first or second on the base classes.

Our method is the only one that reaches the performance of CLIP when it comes to zero-shot performance on unseen classes, whereas all other methods show a drop in performance here, that usually increases with the number of shots, hinting at overfitting.

\subsection{Cross-Dataset Generalization}
In this experiment the models are fine-tuned on ImageNet and then evaluated on the other datasets, thus an improvement on ImageNet (compared to CLIP) is expected. Interestingly both CoCoOp and our approach show an improvement (on average) on the other datasets as well.
Apparently the training samples of ImageNet are numerous and diverse enough to avoid overfitting and the data distribution of ImageNet is closer to the other regarded datasets than the original training set of CLIP. Although our method does not reach the results of CoCoOp on this evaluation we come very close.

\subsection{Domain Generalization}
In this last comparison the models are again fine-tuned on ImageNet and then evaluated on different versions with a clear domain shift.
Here we can see a slight drop of performance between our method and prompt based approaches. This might be due to the fact, that prompt-based approaches only fine-tune the input of the text encoder. As the class names and thus the text encodings are not affected by domain shift, their performance generalizes better. On the other hand we directly update the text and image embedding (and CLIP Adapter only updates the image embedding), which might be problematic as here changes caused by the domain-shift have a more direct effect.

\subsection{Further Analysis}
\paragraph{Ablation} 
\label{pg:ablation}
\begin{table*}
    \centering
    \begin{tabular}{cccccccccc}
\toprule
                & CLIP  & Default   & no cluster& no damp.  & no mask   & Dirac mask    & no reg    & rand. init    & Linear    \\
\midrule
Base            & 69.34 & 83.48     & 83.35     & 83.58     & 83.41     & 83.28         & 83.32     & 82.91         & 81.02     \\
New             & 74.22 & 74.47     & 74.40     & 73.98     & 72.53     & 74.21         & 74.42     & 72.67         & 32.60     \\
Mean            & 71.70 & 78.46     & 78.36     & 78.16     & 77.27     & 78.22         & 78.33     & 76.94         & 45.79     \\
Regularization  & -     & 1.95e-5   & -         & -         & -         & -             & 5e-4      & 3.6e-3        & 6.5e-3    \\
\bottomrule                          
\end{tabular}
    \caption{Ablation of different design decisions in our network. The details of the different experiments are explained in subsection \ref{pg:ablation}}
    \label{tab:ablation}
\end{table*}
Here we discuss the effect different design choices in our approach have on the result. For this we regard the average performance achieved on the Base to New training setup, when using 16 shots (Table \ref{tab:ablation}). 
As already mentioned, it barely makes any difference, whether we use clustering or not ("no cluster"), thus it is a sensible choice to limit the memory requirements. Using the global dampening parameter $\beta$ does lead to an improvement, although it is rather small ("no damp").

Not restricting the interpolation to the training samples ("no mask") leaves the results on the base classes unchanged but leads to overfitting and thus a reduced performance on unseen classes. On the other hand focusing the interpolation exclusively on the training samples ("Dirac mask" equivalent to a $\gamma$ parameter of infinity) leads to the same performance as CLIP for new classes, but of course this way our method cannot improve on unseen classes either, as seen in Tables \ref{tab:cross_dataset} and \ref{tab:domain}. Note that for numerical reasons, we did not actually implement a $\gamma$ parameter of infinity, but clamped $\alpha$ to zero or one, depending on some small distance threshold.

Interestingly leaving out the identity regularization ("no reg") barely has any effect on the results, whereas an initialization to identity seems to be more important ("rand. init"). We assume, that the training does not include enough update steps for effects to be seen. To substantiate this claim we report the actual distance to identity in Table \ref{tab:ablation} (for relevant experiments). Here we can see, that the distance between our adapter and the identity function does correlate with performance. 
Lastly we can see, that only using a single Linear Layer as an adapter without any of our additional improvements leads to significantly worse results, especially on the unseen classes, thus each of our improvements makes only a small difference individually, but together they significantly increase performance.

\paragraph{Training speed}
\begin{figure}
    \centering
    \includegraphics[width=0.45\textwidth]{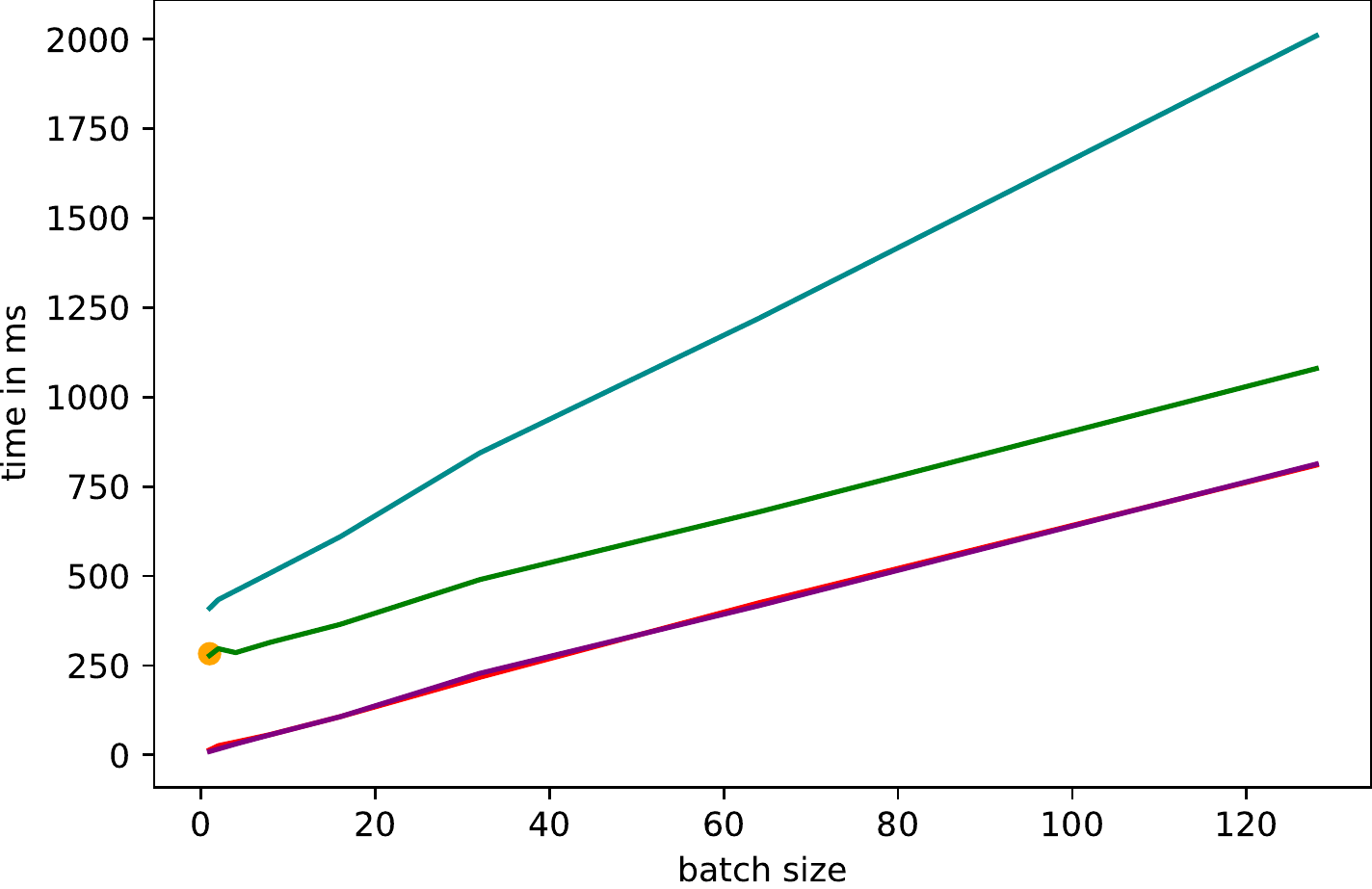}
    \caption{Comparison of timings. Our approach (red), CoOp (green), ProGrad (cyan) and CLIP Adapter (purple). CoCoOp is marked as an orange dot, as a batch size bigger than one does not fit into memory}
    \label{fig:timings}
\end{figure}
A comparison of the training speed between our approach and prompt based methods depends on both the batch size and the number of classes.

As we can precompute the class embeddings, the training time of our method is almost independent of their number. Prompt based approaches instead need to compute the class embeddings in every iteration. On the other hand the number of class embeddings is independent of the batch size, whereas our adapter needs to be applied to every training sample. 
In Figure \ref{fig:timings} we show the timing for a single forward and backward pass depending on the batch size. As can be seen our method and CLIP Adapter are consistently the fastest and the difference in their timings is negligible (it is barely possible to differentiate their lines). Generally the overhead of the computations due to the adapter barely matter, as can be discerned from the similar slope of CoOp and the adapter based methods. 
The number of classes influences the distance between these parallel lines, which signifies the overhead due to the computation of their embedding.

For CoCoOp we only have a single data point, as batches with more than one sample do not fit into memory. Thus, although for a batch size of one the training speed is similar to CoOp, in practice CoCoOp is much slower, as we cannot increase the batch size.
ProGrad is consistently slower than other methods due to additional computations needed for gradient decomposition.

\section{Conclusion}
As the requirements in size, data and compute for state of the art AI models increases, it becomes more and more important to be able use available pre-trained networks for complex downstream tasks. In order to do this we need to be able to fine-tune these models in an efficient manner, preferably without loosing the generalization capability, that makes them so useful in the first place.

We have introduced an extremely simple approach for this task, introducing small linear updates to the embedding space, localized to the datapoints, where we fine-tune. Our model is fast to train and needs a minimal amount of extra parameters, but still reaches state of the art results both on fine-tuned and unseen classes.

In this work we always trained our adapter for optimal performance on a single dataset. A possible future research direction would be to generalize our approach to multiple distinct fine-tuning datasets. It would be possible to use dataset-dependent adapters and interpolation weights, but some further work would be needed to make this scalable.

\paragraph*{Acknowledgements}
This work was funded by the German Research Foundation within the Gottfried Wilhelm Leibniz programme, as well as through the project "Training the Archive" in cooperation with the Ludwig Forum Aachen and the HMKV Hartware MedienKunstVerein, Dortmund. "Training the Archive" is funded by the Digital Culture Programme of the Kulturstiftung des Bundes (German Federal Cultural Foundation). Funded by the Beauftragte der Bundesregierung f\"ur Kultur und Medien (Federal Government Commissioner for Culture and the Media). We would like to thank Dominik Bönisch (heading the "Training the Archive" project) for helpful discussions.

{\small
\bibliographystyle{ieee_fullname}
\bibliography{egbib}
}

\end{document}